\pdfoutput=1

\documentclass[11pt]{article}

\usepackage[T1]{fontenc}
\usepackage[utf8]{inputenc}
\usepackage{times}
\usepackage{latexsym}
\usepackage{microtype}
\usepackage{amsmath}
\usepackage{graphicx}
\usepackage{placeins}
\usepackage{xcolor}
\usepackage{multirow}
\usepackage{booktabs}
\usepackage{verbatim}
\usepackage{fancyvrb}
\usepackage{algorithm}
\usepackage{algorithmic}
\usepackage{boxedminipage}
\usepackage{listings}
\usepackage{inconsolata}
\usepackage{orcidlink}
\usepackage{tabularx}
\usepackage{array}
\usepackage{enumitem}
\usepackage{seqsplit} 

\PassOptionsToPackage{hyphens}{url}
\PassOptionsToPackage{colorlinks=true,urlcolor=blue}{hyperref}

\usepackage[final]{acl}

\title{ttda704 at SemEval-2026 Task 4: Modeling Narrative Structures via Pseudonymization and Multi-View Sentence Alignment}

\author{
\textbf{Tai Tran Tan\textsuperscript{1,2}\thanks{Equal contributions.}\,\orcidlink{0009-0005-4008-6317}},
\textbf{An Dinh Thien\textsuperscript{1,2}\footnotemark[1]\,\orcidlink{0009-0005-4008-6317}}\\
\textsuperscript{1}University of Information Technology, Ho Chi Minh City, Vietnam\\
\textsuperscript{2}Vietnam National University, Ho Chi Minh City, Vietnam\\
\{22521287, 22520010\}@gm.uit.edu.vn
}

\begin{document}
\maketitle
\thispagestyle{plain}
\pagestyle{plain}

\begin{abstract}
    We present our approach to \textit{SemEval 2026 Task 4: Narrative Story Similarity and Narrative Representation Learning}. Our solution uses contrastive learning with fine-tuned sentence transformers to capture narrative similarity across abstract themes, course of action, and outcomes. We develop two pipelines: (Track A) a single-view method that encodes full narratives with smart layer freezing to reduce overfitting, and (Track B) a multi-view method that models theme, plot, and outcome with view-specific projection heads and self-supervised alignment. Both pipelines build on sentence-transformers models and are trained with contrastive loss on synthetic data. The code is available at the following GitHub repository: \href{https://github.com/dinhthienan33/SemEval2026-Task4-ttda704}{https://github.com/dinhthienan33/SemEval2026-Task4-ttda704}.
    \end{abstract}

\section{Introduction}
\label{sec:intro}

Understanding and comparing complex narratives is a fundamental yet persistent challenge in Natural Language Processing. The significance of this problem is highlighted by consecutive SemEval competitions dedicated to narrative analysis ranging from measuring the multidimensional similarity of multilingual news stories \cite{chen-etal-2022-semeval} to characterizing and extracting framing narratives from online media \cite{piskorski-etal-2025-semeval}. Fictional story similarity demands robust representation learning techniques that extend far beyond simple lexical overlap or factual event extraction. To effectively map the architecture of a story, computational models must isolate and encode deep latent dimensions, such as abstract themes, the progression of the plot, and final resolutions \cite{hatzel2024story}.

Historically, handling severe narrative ambiguity has proven extremely difficult. Even advanced generative models frequently falter when faced with scenarios possessing multiple valid interpretations, highlighting the necessity for rigid structural scaffolding to guide model reasoning \cite{keluskar2024can}. Furthermore, attempting to resolve these uncertainties at inference time using computationally heavy mechanisms such as generating multiple reasoning paths via self-consistency voting \cite{wang2023self} introduces massive latency. This makes LLM-based approaches highly impractical for scalable story retrieval, embedding extraction, or large-scale comparative tasks. 

To overcome these limitations, our work shifts away from relying on expensive generative inference at prediction time and instead focuses on efficient narrative representation learning. Inspired by cognitive dual-process theories, which suggest that human cognition adapts its processing depth based on the complexity of the information \cite{kahneman2011thinking}, we use LLMs only in a limited offline role for synthetic data generation and narrative structure extraction, while keeping the final similarity system itself embedding-based and efficient at inference time.

The main contributions of our work are:
\begin{itemize}
    \item{LLM-augmented Narrative Representation.} A deterministic framework that leverages LLMs for high-quality data augmentation and structural extraction, while utilizing fine-tuned sentence transformers for final similarity inference.
    \item{Multi-view structural modeling.} A contrastive learning approach that decomposes stories into theme, course of action, and outcome via view-specific projection heads with self-supervised alignment.
    \item{Robust contrastive training.} Triplet-margin optimization with smart layer freezing to improve generalization while preserving narrative nuance.
\end{itemize}

\section{Related Work}

\subsection{Narrative Representation and Story Embeddings}
While traditional Semantic Textual Similarity (STS) focuses on lexical and sentence-level overlap, narrative similarity requires modeling higher-order structures. This follows the shift in the field towards ``Story Embeddings,'' as formalized by Hatzel et al. \cite{hatzel2024story}, who argue that determining story similarity requires moving beyond keyword matching to capture the underlying narrative graph.
Previous approaches have attempted to model stories hierarchically \cite{lee2020hierarchical}, often focusing on the sequential progression of events. More specific mechanisms, such as re-contextualization through attention \cite{biemann2021narrative}, have been proposed to better encode the flow of a narrative.

\subsection{Sentence Transformers and Backbone Architectures}
To operationalize narrative similarity efficiently, we build upon the Sentence-BERT framework \cite{reimers-gurevych-2019-sentence}, which enables fast and scalable dense text comparison. While foundational contrastive learning approaches \cite{gao2021simcse} excel at general semantic matching, off-the-shelf models are insufficient for fictional texts. Framing narrative comparison as a dense retrieval task requires models to encode deep structural elements rather than mere surface-level semantics \cite{hatzel2024story, younus-qureshi-2025-nlptuducd}. To address this gap and align the representation space with narrative theory, we systematically fine-tune MPNet backbone on the task-provided synthetic narrative triples.

\subsection{Multi-View and Data-Efficient Contrastive Learning}
The separation of narrative similarity into theme, plot, and outcome naturally frames our task as a multi-view learning problem. As demonstrated by Zhang et al. \cite{zhang-etal-2022-multi}, leveraging multiple perspectives of a document significantly improves dense representation compared to standard single-vector compression. We implement this through a shared transformer backbone coupled with aspect-specific projection heads, effectively decoupling complex narrative signals into orthogonal embedding spaces.
Furthermore, the limited availability of human-annotated narrative triples necessitates data-efficient training strategies. To overcome this scarcity and maximize model generalization, we incorporate synthetic data augmentation \cite{jiang-etal-2022-promptbert}.
\section{Task Description}

\textbf{SemEval-2026 Task 4: Narrative Story Similarity and Narrative Representation Learning} challenges participants to develop systems capable of identifying similarities between stories based on their underlying narrative structures rather than surface-level lexical overlaps \cite{semeval2026task4}.

\subsection{Definition of Narrative Similarity}
The task organizers define narrative similarity through \textbf{a specific schema consisting of three core aspects}:

\begin{enumerate}
    \item \textbf{Abstract Theme:} This encompasses the defining constellation of problems, central ideas, and core motifs (e.g., a struggle against nature, a coming-of-age journey).
    \item \textbf{Course of Action:} This describes the sequence of events, conflicts, and turning points. It focuses on the chronological order and the causal links between actions (e.g., a protagonist ignores a warning, suffers a loss, and then recovers).
    \item \textbf{Outcomes:} This refers to the resolution of the plot, such as the fate of the characters or the moral lesson learned. The guidelines emphasize that similar themes can lead to polar opposite outcomes, which must be distinguished.
\end{enumerate}

\subsection{Subtasks}

\begin{itemize}
    \item \textbf{Track A (Triple Classification):} The system is presented with an \textit{Anchor} story and two candidates, \textit{Story A} and \textit{Story B}. The objective is to perform a binary classification to determine which candidate is narratively closer to the Anchor based on the definitions above.
    \item \textbf{Track B (Representation Learning):} The system must generate vector embeddings for individual stories such that the cosine similarity between the vectors reflects their narrative affinity.
\end{itemize}

\section{Methodology}
\label{sec:method}

\begin{figure*}[!t]
\centering
\includegraphics[width=0.8\textwidth]{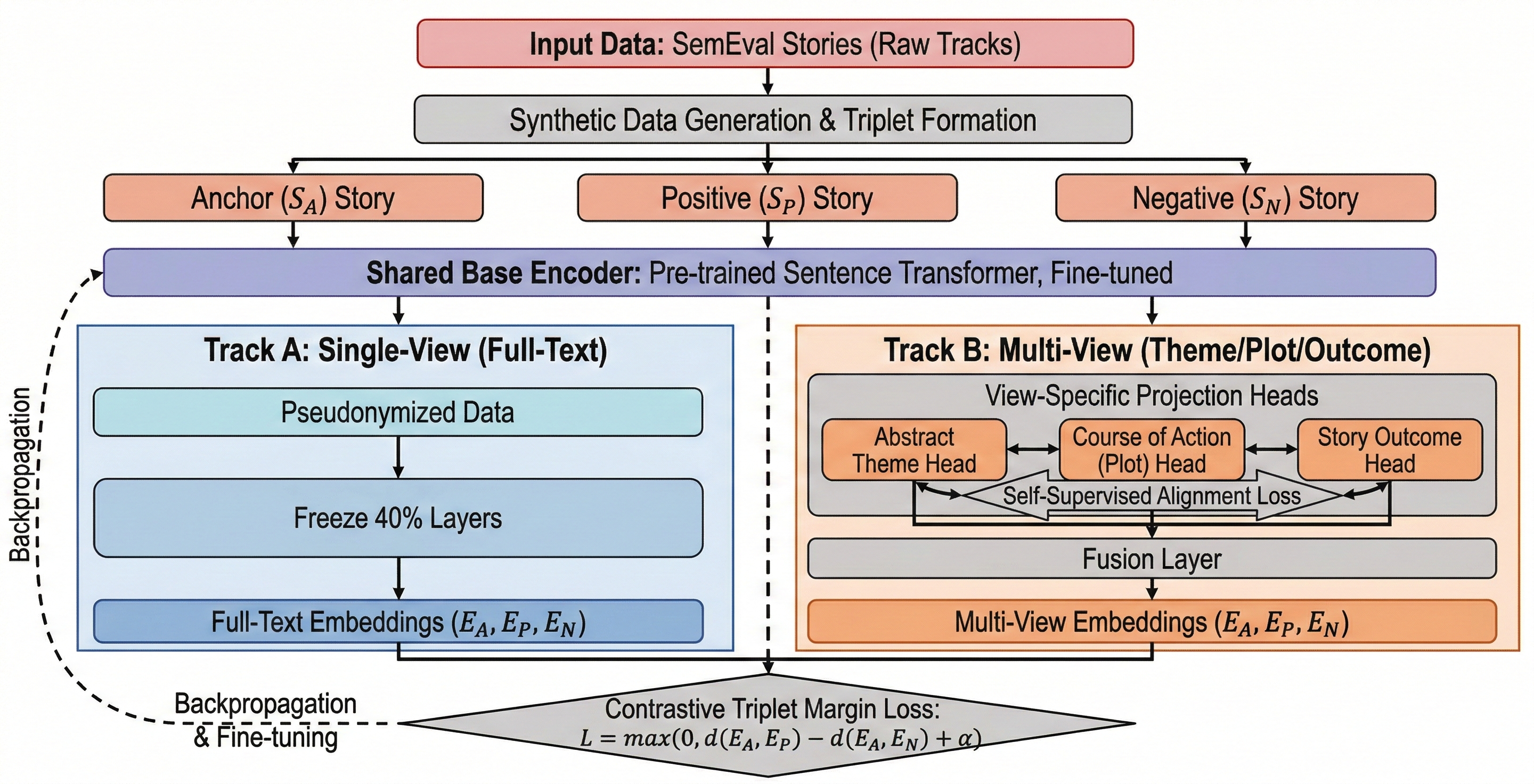}
\caption{High-level overview of the proposed pipeline.}
\label{fig:pipeline_overview}
\end{figure*}

\subsection{Single-View Contrastive Learning for Track A}

Track A requires determining which of two candidate stories is narratively more similar to an anchor story. We employ a single-view contrastive learning approach that learns embeddings directly from full narrative texts, optimized through a pseudonymization step and targeted fine-tuning.

\begin{figure}[!ht]
\centering
\includegraphics[width=0.95\linewidth]{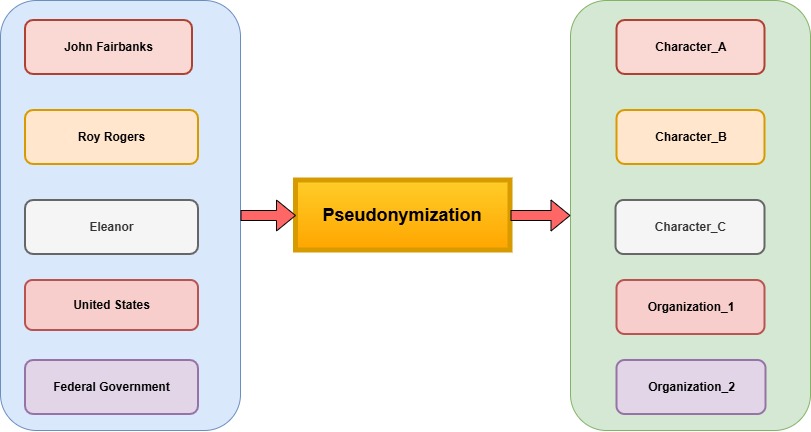}
\caption{Example of the pseudonymization process applied to Track A narratives.}
\label{fig:pseudonymization}
\end{figure}

\textbf{Pseudonymization.} To mitigate surface-level noise and decouple story progression from specific naming conventions, we apply entity-level referential normalization with consistent placeholders while preserving within-story coreference. Repeated mentions of the same character are mapped to the same pseudonym throughout the story, whereas distinct entities receive different placeholders. This makes the transformation identity-preserving at the discourse level, even though lexical content is removed. By pseudonymizing entities into generic tokens (e.g., ``Character\_A'', ``Location\_1''), we reduce the model's reliance on entity overlap---a common source of spurious correlations---and encourage it to focus on narrative structure and semantic progression \cite{hatzel-biemann-2024-tell} (see Figure~\ref{fig:pseudonymization}).

Our scheme is related to, but distinct from, simplified theta-role abstractions in linguistics. Theta roles explicitly encode semantic functions (e.g., agent, patient, experiencer), whereas our preprocessing does not induce a symbolic role inventory; it performs consistent referential normalization and lets the encoder infer event-level relations from context. In implementation, we use a two-stage pipeline with \texttt{fastcoref} and \texttt{spaCy} (\texttt{en\_core\_web\_trf}). We first obtain coreference clusters, infer a cluster type via lightweight voting (\texttt{PERSON}$\rightarrow$\texttt{PER}, \texttt{GPE}/\texttt{LOC}/\texttt{FAC}$\rightarrow$\texttt{LOC}, \texttt{ORG}/\texttt{NORP}$\rightarrow$\texttt{ORG}), and skip clusters without proper-noun evidence to avoid replacing non-entity spans. We then assign deterministic placeholders (e.g., \texttt{Character\_A}, \texttt{Location\_1}, \texttt{Organization\_1}, \texttt{Entity\_1}) using per-type counters and a global mapping dictionary. Next, we run NER fallback on uncovered spans, prevent overlap with a character-level usage mask, and apply deduplicated replacements from right to left (descending character offsets) to preserve index correctness. For each triplet item, one shared mapping is used across \texttt{anchor\_text}, \texttt{text\_a}, \texttt{text\_b}, and \texttt{text} so identical mentions remain consistently pseudonymized within the sample.

\subsubsection{Architecture}
Our model is built upon the \texttt{all-mpnet-base-v2} sentence transformers \cite{song2020mpnet}, which serves as a more robust backbone compared to traditional BERT-based models.

The model produces 768-dimensional embeddings through \textbf{mean pooling} over the output token representations. We deploy this backbone for both tracks. To prevent overfitting and preserve pre-trained semantic knowledge, we implement a \textbf{Smart Layer Freezing} strategy: we freeze all embedding parameters and the bottom 40\% of transformer layers to retain low-level linguistic features, while allowing the top layers to adapt to our task domains without losing generalized language understanding.




\subsection{Multi-View Contrastive Learning for Track B}

\begin{figure*}[!t]
\centering
\includegraphics[width=0.85\textwidth]{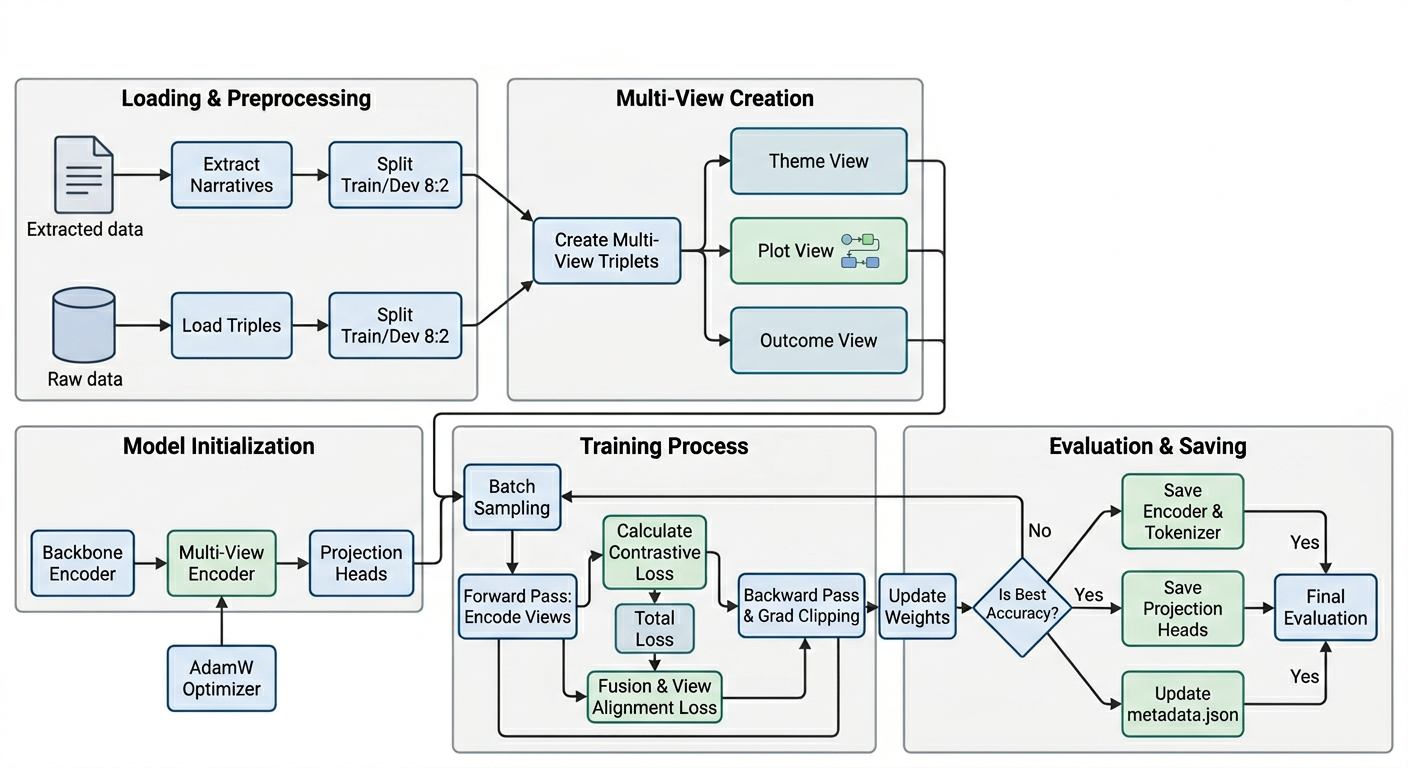}
\caption{Multi-view contrastive learning pipeline.}
\label{fig:multiview}
\end{figure*}

For Track B, we implement a multi-view contrastive learning framework (see Figure~\ref{fig:multiview}) that explicitly models the three narrative dimensions identified in the task annotation guidelines.

\textbf{Narrative Element Extraction.} We decompose each story into three 
distinct views using Large Language Models: (1) \textit{theme}, capturing 
the central controlling idea; (2) \textit{plot events}, representing the 
chronological sequence of state-changing events; and (3) \textit{outcome}, 
describing the final resolution \cite{hobson-etal-2024-story, 
tian2024narratives}. This extraction provides explicit 
structural representations that enable the model to identify semantically 
similar narratives that may differ substantially in surface form 
\cite{li2023claif}.

\textbf{Contrastive Learning.} For each view (theme, plot, outcome), we construct training triplets by matching anchor narratives with their extracted components. This creates a positive pair that shares the same narrative facet and a negative pair drawn from a different story, encouraging view-specific discrimination. The contrastive loss $\mathcal{L}_{\text{con}}$ follows a cross-entropy formulation over scaled cosine similarities \cite{contrastive-learning}:
\begin{equation}
\mathcal{L}_{\text{con}} = -\log \left( \frac{\exp(\frac{\hat{\mathbf{a}} \cdot \hat{\mathbf{p}}}{\tau})}{\exp(\frac{\hat{\mathbf{a}} \cdot \hat{\mathbf{p}}}{\tau}) + \exp(\frac{\hat{\mathbf{a}} \cdot \hat{\mathbf{n}}}{\tau})} \right)
\end{equation}
where $\text{sim}(\mathbf{u}, \mathbf{v}) = \frac{\mathbf{u} \cdot \mathbf{v}}{\|\mathbf{u}\| \|\mathbf{v}\|}$ and $\tau$ is the temperature hyperparameter. A lower $\tau$ sharpens the distribution and increases the penalty for confusing negatives, while a higher $\tau$ smooths similarities and can improve stability.

\textbf{Self-Supervised View Alignment.} To ensure consistency across views with limited data, we introduce an alignment loss $\mathcal{L}_{\text{align}}$. Let
\[
\bar{\mathbf{z}}=\frac{1}{3}\left(\mathbf{z}_{\text{theme}}+\mathbf{z}_{\text{plot}}+\mathbf{z}_{\text{outcome}}\right).
\]
We align the fused embedding $\mathbf{z}_f$ with $\bar{\mathbf{z}}$ via:
\begin{equation}
\begin{aligned}
\mathcal{L}_{\text{align}}
&= \lambda \left\|
\frac{\mathbf{z}_f}{\|\mathbf{z}_f\|_2}
- \frac{\bar{\mathbf{z}}}{\|\bar{\mathbf{z}}\|_2}
\right\|_2^2.
\end{aligned}
\end{equation}
where $\lambda$ is the scaling weight. By minimizing this distance, the model encourages a coherent joint space grounded in all three narrative dimensions.

\section{Results}

\label{sec:results}

In this section, we present the experimental findings for both Track A and Track B. To conduct
a comprehensive evaluation, we employed four distinct state-of-the-art sentence embedding models: the powerful general-purpose \textbf{all-mpnet-base-v2}; the lightweight, distilled \textbf{all-MiniLM-L6-v2} optimized for efficiency; the \textbf{LaBSE} (Language-agnostic BERT Sentence Embedding) model specialized for multilingual bitext retrieval; and \textbf{paraphrase-multilingual-mpnet-base-v2}, a multilingual variant trained on paraphrase data. For brevity, we use \textbf{para-mul-mpnet-v2} to denote the \textit{paraphrase-multilingual-mpnet-base-v2} model in subsequent tables and analysis.

\subsection{Track A Results}

In Track A, our system achieved a private score of \textbf{0.6925}, securing the \textbf{14\textsuperscript{th}} position out of 42 participating teams. For reference, the top-performing system on the leaderboard reached 0.7800. As shown in Tables~\ref{tab:track_a_dev} and Tables~\ref{tab:track_a_test}, pseudonymized data yields consistent improvements. The full Dev-set comparison is provided in Appendix~\ref{appendix:experimental-setup}.


\begin{table}[!ht]
\centering
\small
\renewcommand{\arraystretch}{1.15}
\setlength{\tabcolsep}{5pt}
\begin{tabular}{lcc}
\hline
\textbf{Model} & \textbf{Raw Data} & \textbf{Pseudonymized Data} \\ \hline
all-mpnet-base-v2 & 0.6650 & \textbf{0.6925} \\
para-mul-mpnet-v2 & 0.6625 & 0.6300 \\
LaBSE & 0.6050 & 0.5975 \\
all-MiniLM-L6-v2 & 0.6125 & 0.5800 \\ \hline
\end{tabular}
\caption{Performance comparison of transformer models on Track A (Test set). Bold indicates the best overall score.}
\label{tab:track_a_test}
\end{table}

\paragraph{Performance Analysis.}
The results demonstrate that \textbf{all-mpnet-base-v2} consistently outperforms other models, achieving the highest scores on both Dev and Test sets with pseudonymized data. This aligns with benchmarks on the Massive Text Embedding Benchmark (MTEB) leaderboard, where MPNet's Masked and Permuted Language Modeling (MPLM) objective allows it to capture dependency information more effectively than standard BERT-based models~\cite{song2020mpnet, muennighoff2022mteb}.

\paragraph{Impact of Pseudonymization.}
Pseudonymization yields consistent improvements for the top-performing MPNet model (increasing from 0.6650 to 0.6925 on the Test set). We hypothesize that pseudonymization acts as a normalization step, removing entity-specific noise and forcing the model to focus on the semantic structure of fictional narratives rather than overfitting to specific proper nouns.

\subsection{Track B Results}
For Track B, our team ranked \textbf{6\textsuperscript{th}} out of 25 participants, achieving a private score of \textbf{0.6875} on the Codabench platform (where the top score was 0.7200). Our methodology involved testing various Large Language Models (LLMs): gpt-4o-mini, gpt-4.1, and o3-mini to extract structured schemas from the narratives.
Detailed Dev-set results for Track B are reported in Appendix~\ref{appendix:experimental-setup}.


\begin{table}[!ht]
\centering
\small
\renewcommand{\arraystretch}{1.15}
\setlength{\tabcolsep}{4pt}

\begin{tabularx}{\linewidth}{Xccccc}
\hline
\textbf{Configuration} & $\mathbf{w}_{\text{full}}$ & $\mathbf{w}_{\text{theme}}$ & $\mathbf{w}_{\text{plot}}$ & $\mathbf{w}_{\text{outcome}}$ & \textbf{Dev Score} \\ \hline
Equal weights & 0.25 & 0.25 & 0.25 & 0.25 & 0.625 \\
Full-only & 1.00 & 0.00 & 0.00 & 0.00 & 0.65 \\
Submitted (dev-tuned) & 0.50 & 0.10 & 0.20 & 0.20 & \textbf{0.725} \\
Higher theme & 0.40 & 0.30 & 0.15 & 0.15 & 0.675 \\ \hline
\end{tabularx}

\caption{Fusion-weight ablation on Track B (Dev set). Bold indicates the best overall score.}
\label{tab:track_b_fusion_ablation}
\end{table}


\begin{table}[!ht]
\centering
\small
\renewcommand{\arraystretch}{1.15}
\setlength{\tabcolsep}{5pt}
\begin{tabular}{llc}
\hline
\textbf{Embedding Model} & \textbf{LLM Extractor} & \textbf{Score} \\ \hline
all-mpnet-base-v2 & o3-mini & \textbf{0.6875} \\
all-mpnet-base-v2 & gpt-4.1 & 0.6391 \\
all-mpnet-base-v2 & gpt-4o-mini & 0.6291 \\
LaBSE & o3-mini & 0.6291 \\
LaBSE & gpt-4.1 & 0.6466 \\
LaBSE & gpt-4o-mini & 0.6541 \\
para-mul-mpnet-v2 & o3-mini & 0.5915 \\
para-mul-mpnet-v2 & gpt-4.1 & 0.5940 \\
para-mul-mpnet-v2 & gpt-4o-mini & 0.5614 \\
all-MiniLM-L6-v2 & o3-mini & 0.5414 \\
all-MiniLM-L6-v2 & gpt-4.1 & 0.5664 \\
all-MiniLM-L6-v2 & gpt-4o-mini & 0.5338 \\ \hline
\end{tabular}
\caption{Track B results across different LLM-based schema extractors and embedding models on Track B (Private test set). Bold indicates the best overall score.}
\label{tab:track_b}
\end{table}

\section{Conclusion}
\label{sec:conclusion}

We presented a contrastive learning framework for narrative similarity in SemEval 2026 Task 4, using sentence-transformer embeddings to capture theme, plot progression, and outcomes. Our Track A pipeline relies on single-view full-text embeddings with smart layer freezing, while Track B extends this with multi-view representations and alignment across narrative facets. Across both pseudonymized and raw settings, the approach delivers competitive performance while avoiding expensive generative inference, making it practical for large-scale narrative retrieval and representation learning.

\subsection*{Limitations}
Our models depend on synthetic training data and may not fully generalize to diverse narrative styles or domains without additional adaptation. The multi-view pipeline also relies on automatically extracted narrative elements, which can introduce noise into the learned representations. Due to the shared-task timeline, we were not able to conduct a dedicated manual quality audit of synthetic extractions (e.g., human verification of theme, plot-event coherence, and outcome fidelity), so residual extraction errors may remain unquantified. Finally, fixed fusion weights and temperature settings were tuned on the development data and may not be optimal for unseen domains.

\subsection*{Acknowledgement}
We thank the SemEval-2026 Task 4 organizers \cite{semeval2026task4} for creating the dataset and evaluation framework.

\bibliography{custom}

\appendix

\section{Detailed Experimental Setup}
\label{appendix:experimental-setup}

This appendix provides a comprehensive description of the experimental configurations used for both Track A (narrative similarity classification) and Track B (narrative embedding generation).

\subsection{Reproducibility Checklist}
\begin{itemize}
    \item \textbf{Random seed:} 42 for model training and selection.
    \item \textbf{LLM extraction API/model:} OpenAI Batch API with \texttt{o3-mini} for Track B narrative element extraction.
    \item \textbf{Prompt location:} Full extraction prompt is provided in Appendix~\ref{appendix:prompt-templates} (Table~\ref{tab:prompt-template-v2}).
    \item \textbf{Inference settings (LLM extraction):} temperature $=0.3$, max completion tokens $=2000$, and JSON-only output format.
    \item \textbf{Inference settings (embedding submission):} fixed fusion weights
    $(0.5, 0.1, 0.2, 0.2)$ for (full, theme, plot, outcome), followed by L2 normalization.
    \item \textbf{Hardware:} single NVIDIA GeForce RTX 3090 (24 GB VRAM), detailed in Appendix~\ref{appendix:computational-resources}.
\end{itemize}

\subsection{Data}

All experiments use the officially provided SemEval-2026 Task 4 datasets. The development set contains 200 labeled triples, each consisting of an anchor story, two candidate stories, and a binary label indicating which candidate is narratively closer to the anchor. The test set comprises 400 triples for Track A and 849 individual stories for Track B. Labels for the test set are withheld until the conclusion of the shared task.

For Track A, we construct two data variants from the development set:
\begin{itemize}
    \item \textbf{Pseudonymized}: Stories where person names, organizations, and locations are replaced with standardized references to reduce surface-level matching bias.
    \item \textbf{Raw}: Original story texts without pseudonymization.
\end{itemize}
Both variants use the same 200 development triples for training and evaluation.

For Track B, narrative element extraction is performed on the Track B story corpus using the OpenAI Batch API (see Appendix~\ref{appendix:prompt-templates}). The extracted narrative components are then used to construct multi-view training triplets by matching anchor texts from the Track A development triples with their corresponding extracted theme, plot, and outcome representations.

\subsection{Track A: Single-View Contrastive Learning}

Table~\ref{tab:track-a-config} summarizes the hyperparameter configuration for Track A.

\begin{table}[t]
\centering
\small
\begin{tabularx}{\linewidth}{>{\raggedright\arraybackslash}X l}
\toprule
\textbf{Parameter} & \textbf{Value} \\
\midrule
Base model & \texttt{all-mpnet-base-v2} \\
Embedding dimension & 768 \\
Maximum sequence length & 512 tokens \\
Training epochs & 5 \\
Batch size & 16 \\
Learning rate & $2 \times 10^{-5}$ \\
Weight decay & 0.01 \\
Warmup ratio & 0.1 (linear schedule) \\
Triplet margin ($m$) & 0.3 \\
Layer freezing & 40\% (bottom layers + embeddings) \\
Mixed precision & AMP with GradScaler \\
Random seed & 42 \\
\bottomrule
\end{tabularx}
\caption{Hyperparameter configuration for Track A.}
\label{tab:track-a-config}
\end{table}

Training follows a standard triplet contrastive learning procedure. At each epoch, the full set of training triplets is shuffled and processed in mini-batches. Each triplet is tokenized and encoded in a single forward pass, producing anchor, positive, and negative embeddings. The triplet margin loss is computed using cosine distance, and gradients are scaled via automatic mixed precision. A linear warmup schedule is applied over the first 10\% of total training steps, followed by linear decay.
After each epoch, development accuracy is evaluated by comparing pairwise cosine similarities. The best-performing checkpoint is saved automatically.

\begin{table}[t]
\centering
\small
\renewcommand{\arraystretch}{1.15}
\setlength{\tabcolsep}{5pt}
\begin{tabular}{lcc}
\hline
\textbf{Model} & \textbf{Raw Data} & \textbf{Pseudonymized Data} \\ \hline
all-mpnet-base-v2 & 0.6550 & \textbf{0.6700} \\
para-mul-mpnet-v2 & 0.6400 & \textbf{0.6700} \\
all-MiniLM-L6-v2 & 0.6350 & \textbf{0.6500} \\
LaBSE & \textbf{0.6450} & 0.6350 \\ \hline
\end{tabular}
\caption{Performance comparison of transformer models on Track A (Dev set). Bold indicates the best overall score.}
\label{tab:track_a_dev}
\end{table}

\subsection{Track B: Multi-View Contrastive Learning}

Table~\ref{tab:track-b-config} summarizes the hyperparameter configuration for Track B.

\begin{table}[h]
\centering
\small
\begin{tabular}{ll}
\toprule
\textbf{Parameter} & \textbf{Value} \\
\midrule
Base model & \texttt{all-mpnet-base-v2} \\
Embedding dimension & 768 \\
Projection head hidden dim & 512 \\
Maximum sequence length & 512 tokens \\
Training epochs & 15 \\
Batch size & 32 \\
Learning rate & $2 \times 10^{-5}$ \\
Weight decay & $1 \times 10^{-5}$ \\
Contrastive temperature ($\tau$) & 0.07 \\
Alignment loss weight ($\lambda$) & 0.5 \\
Gradient clipping (max norm) & 1.0 \\
Samples per epoch & 32 \\
Evaluation frequency & Every epoch \\
\bottomrule
\end{tabular}
\caption{Hyperparameter configuration for Track B.}
\label{tab:track-b-config}
\end{table}

The multi-view encoder shares a single transformer backbone across three narrative views. Each view has a dedicated two-layer projection head (768 $\rightarrow$ 512 $\rightarrow$ 768). Training alternates between computing per-view contrastive losses and self-supervised view alignment losses. The total loss is the sum of all six terms (three contrastive + three alignment).

At inference, four embeddings are computed per story (full text, theme, plot, outcome) and fused using fixed weights:
\begin{equation*}
\mathbf{e}_{\text{final}} = 0.5 \cdot \mathbf{e}_{\text{full}} + 0.1 \cdot \mathbf{e}_{\text{theme}} + 0.2 \cdot \mathbf{e}_{\text{plot}} + 0.2 \cdot \mathbf{e}_{\text{outcome}}
\end{equation*}
These fusion weights are selected via manual tuning on the development set rather than analytical optimization. A comparative ablation of alternative weight settings is reported in Section~\ref{sec:results} (Table~\ref{tab:track_b_fusion_ablation}).
The fused embeddings are L2-normalized before submission.

\begin{table}[t]
\centering
\small
\renewcommand{\arraystretch}{1.15}
\setlength{\tabcolsep}{5pt}
\begin{tabular}{llc}
\hline
\textbf{Embedding Model} & \textbf{LLM Extractor} & \textbf{Score} \\ \hline
all-mpnet-base-v2 & o3-mini & 0.6750 \\
all-mpnet-base-v2 & gpt-4.1 & \textbf{0.7250} \\
all-mpnet-base-v2 & gpt-4o-mini & 0.7000 \\
all-MiniLM-L6-v2 & o3-mini & 0.5750 \\
all-MiniLM-L6-v2 & gpt-4.1 & 0.6000 \\
all-MiniLM-L6-v2 & gpt-4o-mini & 0.6000 \\
LaBSE & o3-mini & 0.6250 \\
LaBSE & gpt-4.1 & 0.6500 \\
LaBSE & gpt-4o-mini & 0.6500 \\
para-mul-mpnet-v2 & o3-mini & 0.6750 \\
para-mul-mpnet-v2 & gpt-4.1 & 0.5750 \\
para-mul-mpnet-v2 & gpt-4o-mini & 0.6000 \\ \hline
\end{tabular}
\caption{Track B results across different LLM-based schema extractors and embedding models on Track B (Dev set). Bold indicates the best overall score.}
\label{tab:track_b_dev}
\end{table}

\subsection{Model Selection Across Backbones}

To validate our choice of backbone encoder, we conducted experiments with four sentence transformer models on both tracks under identical hyperparameter settings:

\begin{table}[t]
\centering
\small
\renewcommand{\arraystretch}{1.3} 
\begin{tabularx}{\linewidth}{>{\raggedright\arraybackslash}X}
\toprule
\textbf{Model ID / Backbone} \\
\midrule
\url{sentence-transformers/all-mpnet-base-v2} \\
\url{sentence-transformers/all-MiniLM-L6-v2} \\
\url{sentence-transformers/LaBSE} \\
\url{sentence-transformers/paraphrase-multilingual-mpnet-base-v2} \\
\bottomrule
\end{tabularx}
\caption{Backbone models evaluated for both tracks.}
\label{tab:backbone-models}
\end{table}

All backbone experiments use the same hyperparameters, data splits, and random seeds. The final submission for both tracks uses \texttt{all-mpnet-base-v2}, selected based on development set performance.

\section{Prompt Templates}

\label{appendix:prompt-templates}
This appendix presents the prompt template used for narrative element extraction in the Track B pipeline. Narrative components are extracted from each story via the OpenAI Batch API using the \texttt{o3-mini} model with a temperature of 0.3 and a maximum of 2,000 completion tokens. The response format is constrained to JSON output. The extracted components: theme, plot events, and outcome are subsequently used to construct multi-view training triplets and to generate view-specific embeddings at inference time.
The prompt design is grounded in computational narratology principles, specifically the notions of Tellability and Changes of State. Events are selected based on whether they cause physical, mental, or social modifications and whether they produce affective fluctuation or alter character relationships.
 
\begin{table*}[!t] 
\normalsize
\centering
\renewcommand{\arraystretch}{1.12}
\begin{tabularx}{\textwidth}{|X|}
\hline
\textbf{System Prompt \& Instructions} \\
\hline
\texttt{You are an expert in Computational Narratology and Narrative Analysis. Your task is to extract the core narrative structure from the story summary below, focusing on "Tellability" and "Changes of State."} \\
\\
\textbf{STORY SUMMARY:} \\
\texttt{\{story\_summary\}} \\
\\
\textbf{EXTRACTION INSTRUCTIONS:} \\
\begin{itemize}[leftmargin=*, nosep]
    \item \textbf{Theme} (1-3 sentences): Identify the central controlling idea or semantic abstract of the story. Focus on the underlying motivation or the specific "script" (e.g., revenge, redemption, sacrifice) that governs the narrative logic.
    \item \textbf{Plot Events} (5-10 key events): Extract the main chronological sequence of events.
    \begin{itemize}[leftmargin=1.5em, nosep]
        \item \textit{Change of State:} Only select events that cause a physical, mental, or social modification in the characters or the world.
        \item \textit{Affective Impact:} Prioritize events that create "Affective Fluctuation" (shifts in tension) or alter the relationships between characters.
        \item \textit{Structure:} Phrase each event concisely as [Subject] + [Predicate/Action] + [Object/Outcome].
    \end{itemize}
    \item \textbf{Outcome} (1-2 sentences): Describe the final "Resolution State." Detail how the character network or the narrative world has fundamentally changed compared to the beginning.
\end{itemize}
\\
\textbf{RESPONSE FORMAT (JSON):} \\
\texttt{\{} \\
\texttt{\ \ "theme": "...",} \\
\texttt{\ \ "plot\_events": ["Event 1", "Event 2", "..."],} \\
\texttt{\ \ "outcome": "..."} \\
\texttt{\}} \\
\\
\textbf{IMPORTANT:} Filter for High Tellability. Ensure strict chronological order. Return ONLY the JSON object. \\
\hline
\end{tabularx}
\caption{Prompt template for narrative element extraction used in Track B. The template instructs the model to decompose each story into theme, plot events, and outcome following computational narratology criteria.}
\label{tab:prompt-template-v2}
\end{table*}

\section{Computational Resources}
\label{appendix:computational-resources}

All training and inference experiments were conducted on a single NVIDIA GeForce RTX 3090 GPU with 24 GB VRAM.

Track A training uses automatic mixed precision (AMP) with gradient scaling, which reduces both memory consumption and wall-clock time. Track B training processes batches of multi-view triplets with periodic CUDA cache clearing to manage GPU memory. Both tracks perform evaluation after each epoch, which is included in the reported training times.

\end{document}